\documentclass[conference]{IEEEtran}
\IEEEoverridecommandlockouts

\usepackage{cite}
\usepackage{amsmath,amssymb,amsfonts}
\usepackage{algorithmic}
\usepackage{graphicx}
\usepackage{textcomp}
\usepackage{xcolor}
\usepackage{url}
\usepackage{hyperref}
\usepackage{cleveref}
\usepackage{balance}
\def\BibTeX{{\rm B\kern-.05em{\sc i\kern-.025em b}\kern-.08em
    T\kern-.1667em\lower.7ex\hbox{E}\kern-.125emX}}

\begin{document}

\title{Automatic Extraction of Time-windowed ROS Computation Graphs from ROS Bag Files}

\author{
 \IEEEauthorblockN{Zhuojun Chen}
 \IEEEauthorblockA{
\textit{Vrije Universiteit Amsterdam}\\
 Amsterdam, The Netherlands \\
 \href{mailto:berry.chen@student.vu.nl}{z6.chen@student.vu.nl}
}
\and
\IEEEauthorblockN{Michel Albonico}
\IEEEauthorblockA{
\textit{Federal University of Technology, Paran\'a (UTFPR)}\\
 Francisco Beltr\~ao, Brazil \\
\href{mailto:michelalbonico@utfpr.edu.br}{michelalbonico@utfpr.edu.br}
}
 \and
 \IEEEauthorblockN{Ivano Malavolta}
 \IEEEauthorblockA{
\textit{Vrije Universiteit Amsterdam}\\
 Amsterdam, The Netherlands \\
 \href{mailto:i.malavolta@vu.nl}{i.malavolta@vu.nl}
}
}

\maketitle

\begin{abstract}
Robotic systems react to different environmental stimuli, potentially resulting in the dynamic reconfiguration of the software controlling such systems.
One effect of such dynamism is the reconfiguration of the software architecture reconfiguration of the system at runtime. Such reconfigurations might severaly impact runtime properties of robotic systems, e.g., in terms of performance and energy efficiency.
The ROS \emph{rosbag} package enables developers to record and store timestamped data related to the execution of robotic missions, implicitly containing relevant information about the architecture of the monitored system during its execution. 
In this study we discuss about our approach for statically extracting (time-windowed) architectural information from ROS bag files. The proposed approach can support the robotics community in better discussing and reasoning on the software architecture (and its runtime reconfigurations) of ROS-based systems.
We evaluate our approach against hundreds of ROS bag files systematically mined from 4,434 public GitHub repositories.
\end{abstract}


\section{Introduction}
% This document is a model and instructions for \LaTeX.
% Please observe the conference page limits. 

Robotic software has dynamic demands over its runtime execution.
Let us consider a multi-robot warehouse environment (RWARE)~\cite{BOYSEN2019396} as a motivation, where a robot may go through a factory plan by using a navigation/localization stack and an already-known map.
However, at a certain point, a person, another robot, or an object may be in/cross its way, when it needs to react and plan a new path~\cite{han2019}.
Finally, when it finishes the current task, another task must be assigned to the robot~\cite{bolu2021}.

The dynamism of robotic missions may result in constant software architecture reconfiguration, with volatile nodes and messaging activities varying over time.
Automatically extracting such an underlying software architecture reconfiguration
helps researchers and practitioners to assess their ROS-based systems in different ways, potentially enabling them to identify the need for a redesign of the system.
One could use this information to debug errors, assess whether a certain ROS node is either under- or over-used, or even assessing the energy consumption of ROS-based systems.
%Furthermore, such information could help indicate that software artifacts are loaded in a way it could waste both, processing power and/or energy.
However, \textbf{gathering architectural information at run-time is not trivial}, underlying difficulties such as deploying a complete robotic system.
Furthermore, existing static approaches only allow getting the final architecture, without inspecting its reconfiguration during the mission execution~\cite{corke2015, Duan2023, lu2020, santos2019haros}. 

In this study we consider ROS~\cite{ros}
%\footnote{\url{https://www.ros.org/}}
as the robotic framework, the de facto standard for developing and prototyping robotic systems. 
The ROS ecosystem includes a file format called \emph{bag}~\cite{bags},
%\footnote{\url{http://wiki.ros.org/Bags}}
 which is able to represent the messages exchanged by various ROS nodes at runtime; the tool for recording and reading ROS bag files is called \emph{rosbag}~\cite{rosbag}.
%\footnote{\url{http://wiki.ros.org/rosbag}}.
Those bag files contain rich architectural information, such as the nodes that publish and subscribe in/to certain ROS \emph{topics}, and the messaging timestamp, which can help in reconstructing the dynamic ROS architecture reconfiguration.

We propose a three phases approach, which i) extracts ROS nodes and topic communication from ROS bag files, ii) slices such information according to a user-defined time window, and iii) generates the computation graph of the sliced architectural information within the considered time window. 
Besides the construction of time-windowed computation graphs, our approach computes the average frequency of each topic communication, potentially highlighting the intensity of the usage of each ROS node within the system.

We evaluated the approach among 1,373 ROS bag files (for both, ROS 1 and ROS 2) gathered from GitHub repositories and large ROS bag files ($\approx50MB$) from a public ROS project.
Our approach is able to generate computation graphs in both cases, only missing potentially-relevant nodes for 49 ROS bag files, which can be obtained from other sources, e.g., source code, ROS launch files, etc.  %when our approach could still be used as a complement given its innovative time-windowed and communication frequency features.

\begin{figure}
    \centering
    \includegraphics[width=0.45\textwidth]{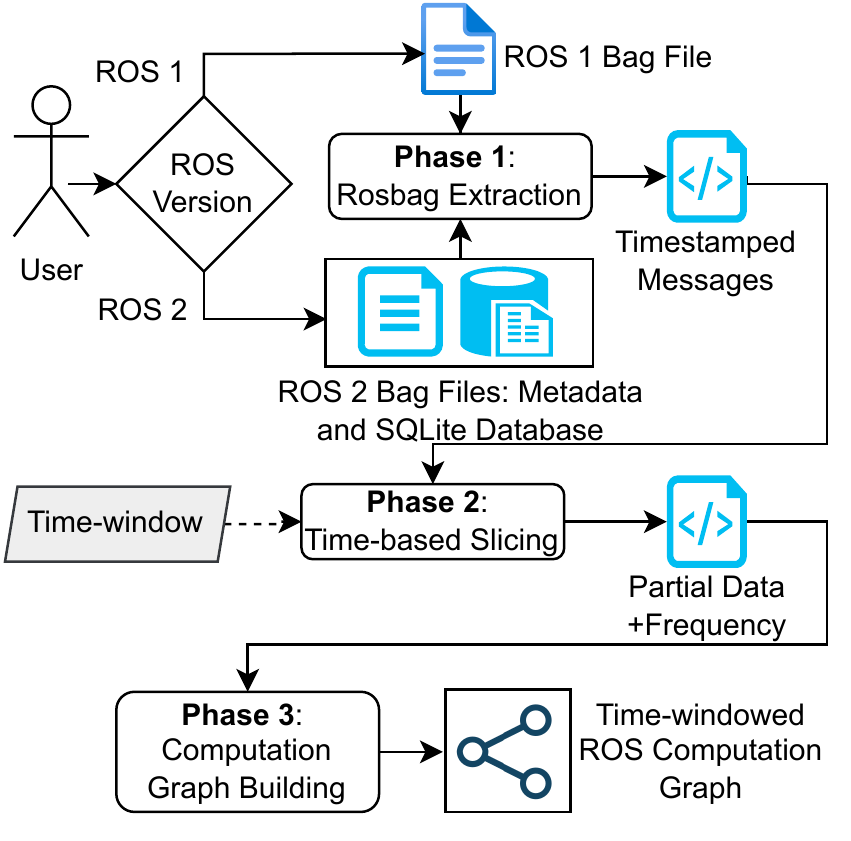}
    \caption{The workflow of the proposed approach.}
    \label{fig:approach}
\end{figure}

\section{Proposed Approach}
\label{sec:approach}

As shown in \Cref{fig:approach}, our approach is composed of three main phases: i) rosbag extraction, ii) time-window slicing, and iii) computation graph building. %, which is explained with more details in the sequence.

\noindent
1) \textbf{Rosbag Extraction}: in this phase, we read the ROS bag content data and store it in a tabular format (CSV file), which is broadly used for data conversion and analysis.
Even though ROS 1 and ROS 2 have different bag formats, both of them are supported by the approach. 
%While ROS 1 relies on a single bag file, ROS 2 counts on a metadata file and a \emph{SQLite} database.
Both bag formats contain key information, such as timestamped topic messages and nodes subscription, which can be used to reconstruct the ROS computational graph.
For extracting the ROS bag file data, we rely on \emph{rosbags} Python library~\cite{rosbags}.
%\footnote{\url{https://pypi.org/project/rosbags/}}]
%, which is able to deserialize both, ROS 1 and ROS 2 bag files without further modification.

\noindent
2) \textbf{Time-window Slicing}: in this phase, we select only the computation graph components within a time interval (time-window passed as a parameter),
%and the frequency of such message exchanging in that interval.
which benefits from the data that is already tabulated in the CSV file from phase 1
%, and therefore, a simple selection is performed.
%If no argument is passed, all the nodes and topics are considered.

\noindent
3) \textbf{Computation Graph Building}: finally, in this phase, we generate a computation graph compatible with RQT~\cite{rqtgraph}, 
which is a standard among ROS community.
For graph generation, we rely on Graphviz graph visualization tool~\cite{graphviz}, 
%\footnote{\url{https://graphviz.org/}}
which is fed by a Python script that reads all the registers from the CSV file and creates the graph accordingly.

\section{Preliminary Results}

For the purpose of evaluating the ability of our approach to deal with different types of ROS bag files, we use it among 1,374 public ROS bag files extracted from 4,434 GitHub repositories.
%by using the GitHub REST API~\cite{gitrest}
%with \textbf{rosbag} as search keyword~\cite{albonico_sbcars} in the title, description, or commit across all repositories with the GitHub platform. 
%This results in 4,434 repositories to be analyzed.
%By automating a search in those GitHub repositories with Selenium~\cite{selenium}, we found 1,341 ROS 1 bag files in 414 repositories.
%We also manually searched for ROS Bag 2 files, in order to evaluate our approach 
%only found a few ROS 2 bag files on GitHub, we also obtained 1 extra file by asking the ROS community via the ROS Discourse forum~\cite{rosdiscourse}, other 3 by manually crawling GitHub with alternative terms (such as 'ROS2 + data'), and we finally generated another one with a simple Turtlesim execution~\cite{turtlesim}.
%The main reason for the small number of bag files is that such files tend to be considerably large, depending on the robot mission, and therefore, cannot be stored in GitHub.
%However, we gather a list which is num from different purposes of files for the evaluation of our work.
%Basically, Selenium opens the repository URL and search form bag extensions, listing the link to files that are found.
%We make available a spreadsheet containing the link to all the found ROS bag files \textbf{[REF]}.
Among those files, our approach could not create a complete computation graph for 238 of them since they missed \emph{rosout} topic (used for logging); and as a consequence, there was no reference to ROS nodes.
However, even for those files, we were able to get topics' information, including messaging frequency.
%We make all the scripts and links to the found bag files available in a replication package\footnote{\url{https://github.com/S2-group/icra-ws-robotics-rosbag}}.
%\todo[inline]{Pct. of repos and files that are not compatible - it could be a take away.}

We also evaluated the scalability of our approach while extracting the architectural information of the Mars Emulation Terrain (MET) project~\cite{met}
%\footnote{\url{https://starslab.ca/enav-planetary-dataset/}}
, which makes available very large ROS bag files (up to $\approx50$ GB).
For the largest file, without a time window (i.e., considering the overall architecture), the approach takes $\approx 8$ minutes to complete.
Its timestamping indicates it corresponds to an 11.8 minutes robotic mission. 
When a time window of 1 minute (at the beginning) was considered, it took less than 1 minute to complete, where the time increases by only a few milliseconds when we move the 1-minute time window to the end.
In all the cases, the approach was able to generate a consistent computation graph\footnote{We do not have a reference of what would be a correct graph, but nodes and topics are compatible with what is described in the project Web page.}. 

While analyzing the data, we observe that in a great part of ROS bag files, topic message frequency is skewed along execution time, which makes sense since the robotic system dynamics.
%ROS communication is not constant.
We also perceive that topics have a discrepancy in the overall average frequency.
This is also expected since topics such as \emph{rostout} (logging mechanism), receive messages constantly, while others, such as object avoidance, are more event-responsive.
Such perceptions strengthen our conjecture that ROS architecture is highly dynamic over time, and should be covered so the community is able to better understand potentially related issues, such as low performance and energy inefficiency.

The interested reader can refer to our replication package\footnote{\url{https://github.com/S2-group/icra-ws-robotics-rosbag}} to (i) verify and reuse the current implementation of the proposed approach and (ii) verify the ROS bags dataset used for the evaluation.

\section{Limitations}

ROS bag files do not contain \textit{all} possible information about the communication graph of a ROS-based system. For example, in some cases ROS nodes may publish/subscribe to different topics depending on the environment, or ROS services are not present at all in the bag files format.
%with the timestamped messages from/to the topics that nodes in launch files publish/are subscript, it is possible to infer when they should be fired.
%The 
Our approach reaches the goals of getting time-windowed and statistical information that represents consistent architectural reconfiguration.
%, and extra nodes information could be covered by other existing approaches.
Nodes information can be easily obtained from other sources, such as ROS \emph{launch files} and source code, even using existing approaches, when our approach would still help as a complement.

We only aimed at extracting ROS-specific information, without considering interoperability with other systems.
It is important that future work addresses the integration with RQT tool, which is largely used by the ROS community.
%For instance, it could be a way to either load the architectural information file or an RQT plugin.
It is also necessary to implement a feature that enables slicing per period of time, such as at each minute, where it would be possible to see the architectural evolution in a framed sequence.

\section{Conclusion}

This paper's approach helps in statically extracting ROS architectural reconfiguration from ROS bag files.
%This is done by extracting the timestamped data in those files according to a preset time-window argument.
It worked accordingly against hundreds of ROS bag files publicly available, which also include very large files ($\approx 50MB$).
%Our approach is made publicly available by means of a replication approach, 
All the approach' artifacts are publicly available, which may benefit the community in debugging ROS systems with a fine-grained inspection of their architectural reconfiguration during a mission.
The replication package also includes the list of public ROS bag files, which can also further projects.
Our main goals for the future are to integrate our approach with others that enable extracting nodes information hidden from ROS bag files, and 
%, from other sources such as source code and launch files.
to improve the approach features according to the reported in the limitations section.

% \section*{Acknowledgment}
% The preferred spelling of the word ``acknowledgment'' in America is without 
% an ``e'' after the ``g''. Avoid the stilted expression ``one of us (R. B. 
% G.) thanks $\ldots$''. Instead, try ``R. B. G. thanks$\ldots$''. Put sponsor 
% acknowledgments in the unnumbered footnote on the first page.

% 

% \clearpage

\balance

\bibliographystyle{plain}
\bibliography{bibliography}

\end{document}